\documentclass[10pt,twocolumn,letterpaper]{article}

\usepackage{cvpr}              

\usepackage[accsupp]{axessibility}

\newcommand\blfootnote[1]{%
  \begingroup
  \renewcommand\thefootnote{}\footnote{#1}%
  \addtocounter{footnote}{-1}%
  \endgroup
}

\definecolor{cvprblue}{rgb}{0.21,0.49,0.74}
\usepackage[pagebackref,breaklinks,colorlinks,allcolors=cvprblue]{hyperref}

\def\paperID{6181} 
\def\confName{CVPR}
\def\confYear{2026}

\def\eg{\emph{e.g.}}
\def\etal{{\em et al.}}
\usepackage{overpic}
\usepackage{multicol,multirow}
\usepackage{rotating}
\usepackage{colortbl} 
\usepackage[table]{xcolor}

\title{SemiGDA: Generative Dual-distribution Alignment for Semi-Supervised Medical Image Segmentation}

\author{
    Kaiwen Huang$^{1}$ \quad 
    Yi Zhou$^{2}$ \quad 
    Yizhe Zhang$^{1}$ \quad 
    Jingxiong Li$^{1}$ \quad 
    Tao Zhou$^{1*}$ \quad 
    \\
    $^{1}$Nanjing University of Science and Technology, Nanjing, China\\
    $^{2}$Southeast University, Nanjing, China \\
    {\tt\small kwhuang@njust.edu.cn, taozhou.ai@gmail.com}
}

\begin{document}
\maketitle

\blfootnote{$^*$Corresponding author.}

\begin{abstract}
Semi-supervised learning addresses label scarcity and high annotation costs in medical image segmentation by exploiting the latent information in unlabeled data to enhance model performance. Traditional discriminative segmentation relies on segmentation masks, neglecting feature-level distribution constraints. This limits robust semantic representation learning and adaptive modeling of unlabeled data in scenarios with few labels.
To address these limitations, we propose SemiGDA, a novel Generative Dual-distribution Alignment framework for semi-supervised medical image segmentation. Our SemiGDA overcomes the reliance of discriminative methods on large labeled datasets by aligning feature and semantic distributions to boost semantic learning and scene adaptability. Specifically, we propose a Dual-distribution Alignment Module (DAM), which employs two structurally distinct encoders to model image and mask feature distributions. It enforces their alignment in the latent space via distributional constraints, establishing structured feature consistency. Moreover, we design a Consistency-Driven Skip Adapter (CDSA) strategy, which introduces dual skip adapters (Image and Mask) to fuse multi-scale features via skip connections. Using a consistency loss, CDSA enhances cross-branch semantic alignment and reinforces fine-grained semantic consistency.
Experimental results on diverse medical datasets show that our method outperforms other state-of-the-art semi-supervised segmentation methods. 
\textbf{Code is released at: https://github.com/taozh2017/SemiGDA}. 

\end{abstract}

\section{Introduction}
\label{sec:intro}

\begin{figure}[!t]
	\centering
	\begin{overpic}[width=0.48\textwidth]{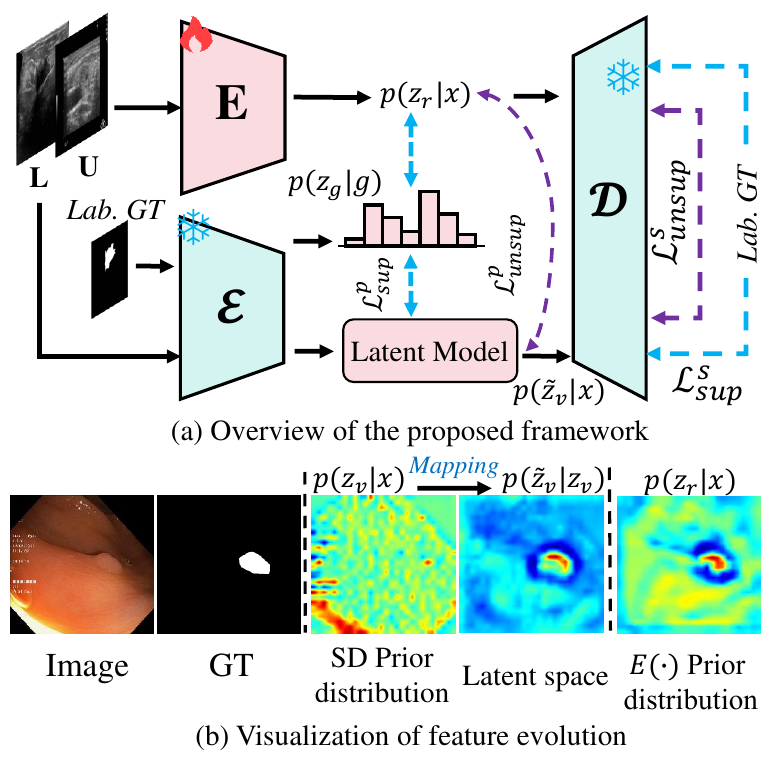}
    \end{overpic}
    \vspace{-0.70cm}
	\caption{{
    Illustration of the proposed SemiGDA. (a) Overview of the prior distribution-based semi-supervised architecture. By utilizing feature-level mask constraints and generative segmentation, the model offers high robustness to limited hard labels. (b) Feature evolution: The Latent Mapping Model aligns the Stable Diffusion (SD) prior to a valid manifold, preventing feature collapse and ensuring stable training.
 }}
 \vspace{-0.40cm}
    \label{fig:Fig1}
\end{figure}

Medical image segmentation plays a crucial role in disease diagnosis, treatment planning, and surgical guidance~\cite{mei2025survey, jiao2023learning}. 
However, its fully-supervised paradigms~\cite{ronneberger2015u, cao2022swin, liu2022convnet} require large, expertly-annotated datasets, which are costly and difficult to acquire. 
Semi-supervised Medical Image Segmentation (SMIS)~\cite{yu2019uncertainty, chen2021semi, huang2025learnable} has thus emerged as a promising alternative, leveraging both limited labeled data and abundant unlabeled data to reduce annotation dependency while achieving robust performance. 


Existing SMIS methods are primarily based on pseudo-labeling and consistency learning. Pseudo-labeling methods~\cite{zeng2023ss, yao2022enhancing} generate labels for unlabeled data and iteratively refine the model, but their performance can be hampered by noisy initial predictions, which affects training stability. In contrast, consistency learning methods~\cite{huang2025uncertainty, yang2023revisiting} improve model generalization by enforcing output invariance under different perturbations applied to inputs or features. 
The representative Mean Teacher framework~\cite{tarvainen2017mean} enforces consistency between teacher and student models. 
Moreover, some methods~\cite{wang2023mcf, xu2023dual, gao2023correlation} utilize dual-stream networks to capture different data features, employing mutual learning by enforcing cognitive consistency across the same input. However, these approaches are primarily based on the discriminative paradigm, focusing on per-pixel classification, which restricts their robustness, particularly in semi-supervised tasks with limited labeled data. Furthermore, when addressing complex segmentation tasks, they often fail to effectively capture image structures and integrate contextual semantic information.


To address these limitations, generative models offer a promising alternative. 
Representative generative models, including Generative Adversarial Networks (GANs)~\cite{goodfellow2014generative} and Variational Autoencoders (VAEs)~\cite{kingma2013auto}, have been widely adopted for data augmentation. 
In semi-supervised tasks, these generative methods are widely applied to use adversarial mechanisms to make the prediction distribution of unlabeled images as close as possible to that of labeled images~\cite{chen2021mtans, peiris2021duo}. These methods typically involve a discriminator to differentiate between the predictions of labeled and unlabeled data, thus optimizing the generative model's performance~\cite{he2023bilateral}. However, such adversarial training often faces convergence challenges~\cite{jiao2023learning}.

To this end, we propose a novel Generative Dual-distribution Alignment framework for semi-supervised medical image segmentation (SemiGDA). Its core innovation is to map images and masks into a latent space and align their prior distributions, leveraging these aligned variables to synthesize high-quality segmentation masks with limited labeled data. To enable generative segmentation in semi-supervised scenarios, we propose a Dual-distribution Alignment Module (DAM), which involves a dual-encoder structure to enforce consistency on the same input, as illustrated in~\cref{fig:Fig1}(a). Image features are then translated into mask features via a learnable encoder and a latent space mapping module, effectively aligning the prior distribution to a valid manifold, as shown in~\cref{fig:Fig1}(b). Furthermore, to overcome the generative decoder's limitations in capturing global context and semantic detail, we present a Consistency-Driven Skip Adapter (CDSA) strategy integrating multi-scale adapters during decoding to enhance robustness and semantic alignment. Finally, an Annotation Conversion and Reversion (ACR) strategy is designed to resolve dimensionality mismatches with ground truth, ensuring compatibility while preserving semantic integrity.

To summarize, our main contributions are four-fold: 
\begin{itemize}


\item To the best of our knowledge, the proposed SemiGDA is the first work to directly address the overfitting and poor generalization of discriminative models under limited annotations through a generative paradigm with dual-distribution alignment.

\item We present the dual-distribution alignment module to model the distribution transformation between images and masks, facilitating effective alignment of image and mask feature distributions in a latent space.

\item A consistency-driven skip adapter is proposed to enhance the decoder's capacity for capturing semantic details and improving robustness, enabling the generative model to be better adapted to the segmentation task.


\item Experiments results on four distinct medical benchmarks demonstrate that the proposed method outperforms existing state-of-the-art SMIS approaches.

\end{itemize}

\begin{figure*}[!t]
	\centering
	\begin{overpic}[width=0.99\textwidth]{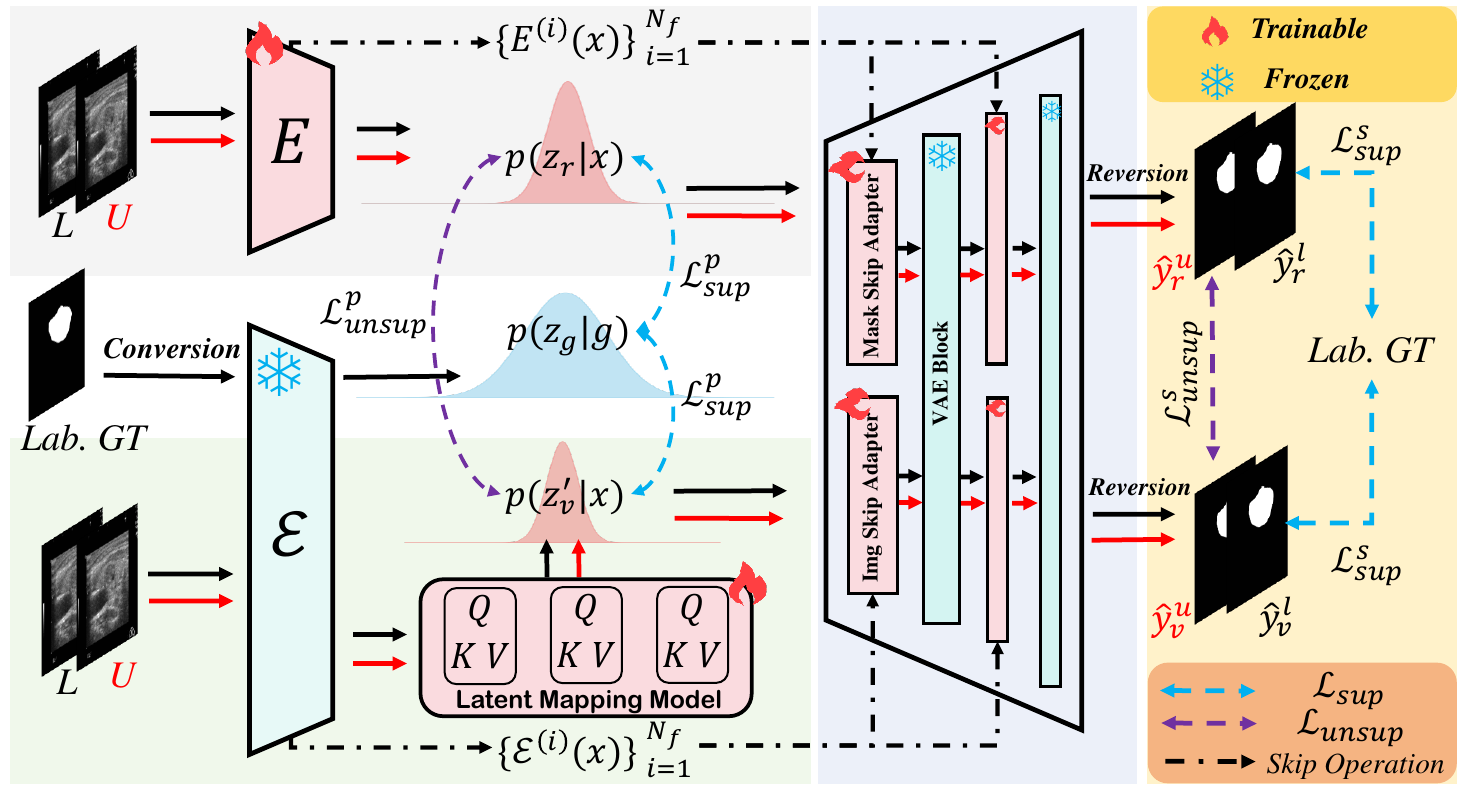}
    \end{overpic}\vspace{-0.0cm}
	\caption{{Overview of the proposed framework.
    The trainable components include a trainable encoder~(\ie~ResNet as backbone), a mapping network, a lightweight skip adapter module, and a VAE network with frozen parameters. In the first stage, the encoder and mapping network align the input image with the label's prior distribution. In the second stage, the transformed features are passed to the VAE decoder. The skip adapter module is added to improve model performance within the segmentation framework. Additionally, label transformations and inverse transformations are applied to ensure the ground truth conforms to VAE input requirements.
 }}\vspace{-0.1cm}
    \label{fig:Model}
\end{figure*}

\section{Related Works}

\subsection{Semi-supervised Medical Image Segmentation}

Semi-supervised medical image segmentation leverages a small amount of labeled data and a large volume of unlabeled data to train models, enhancing segmentation performance by utilizing the potential of unlabeled data~\cite{jiao2023learning}. 
Traditional approaches to semi-supervised medical image segmentation predominantly rely on pseudo-labeling~\cite{jiang2024ph, zhang2022discriminative, li2023segment} and consistency learning~\cite{wang2023mcf, basak2022exceedingly, shu2022cross, yang2023revisiting}. 
Pseudo-labeling uses an initially trained model to predict labels for unlabeled data, generating pseudo-labels that are iteratively refined to improve model accuracy. 
zeng~\etal~\cite{zeng2023ss} proposed a Semi-Supervised Tri-Branch Network (SS-TBN) that combines segmentation and classification tasks with pseudo labeling and inter-slice consistency to improve COVID-19 diagnosis with limited annotated data.
Consistency learning, widely used in semi-supervised learning, is exemplified by the Mean Teacher (MT)~\cite{tarvainen2017mean} model, where the teacher model's parameters are updated via Exponential Moving Average (EMA), and the student model is optimized by minimizing the output discrepancy between the teacher and student networks.
However, the student model is often influenced by the cumulative errors of the teacher model, leading to the development of various derivatives of the teacher-student framework~\cite{yu2019uncertainty, zhao2024alternate, huang2025text}.


\subsection{Generative Models for Image Processing}

In medical image processing, generative models have demonstrated great potential in generating synthetic images, improving resolution, and reconstructing damaged areas~\cite{dayarathna2024deep}.
In scenarios with limited medical annotations, many methods~\cite{zhang2025generative, ye2023synthetic} expand the dataset for downstream tasks by generating new data using generative models.
Specifically, VAE~\cite{kingma2013auto} can achieve efficient image generation and reconstruction through variational inference, but faces limitations in detail and diversity.
Generative Adversarial Networks (GANs)~\cite{goodfellow2014generative} learn the data's latent distribution and generate realistic high-dimensional samples through adversarial training, but suffer from issues like mode collapse and vanishing gradients during training.
Meanwhile, the generative approach based on diffusion models trained on large-scale datasets, such as Stable Diffusion~\cite{rombach2022high}, alleviates the issue of data scarcity~\cite{zhang2025generative} and has yielded promising results. However, the iterative training process is complex and time-consuming.
Currently, the direct application of generative models to segmentation research is quite limited.
GMS~\cite{huo2025generative} utilizes a pre-trained vision model to extract latent representations and learn a mapping function to generate segmentation masks, improving generalization and reducing overfitting.
GSS~\cite{chen2023generative} frames semantic segmentation as a mask generation problem by modeling the posterior distribution of segmentation masks and using a conditioning network to optimize the divergence between mask and image priors.
GMMSeg~\cite{liang2022gmmseg} combines generative and discriminative models by using Gaussian Mixture Models (GMMs) for class-conditional densities and end-to-end discriminative training.

\section{Proposed Method}



\textbf{Setting and Overview}. In the context of semi-supervised learning, the training dataset consists of labeled data $\mathcal{D}^L=\{(x^l_i,y^l_i)\}_{i=1}^{N_l}$ with $N_l$ samples and unlabeled data $\mathcal{D}^U=\{x^u_i\}_{i=N_l+1}^{N_l+N_u}$ with $N_u$ samples, where $N_l \ll N_u$. Here, $x_i \in \mathbb{R}^{H \times W \times 3}$ denotes an input image, and $y_i \in \{0,1\}^{K \times H \times W}$ represents the corresponding ground-truth annotation with $K$ classes.

\cref{fig:Model} illustrates the overall pipeline of the proposed method. First, both labeled and unlabeled data are input into a frozen VAE encoder. The outputs of the VAE encoder are then passed into the latent mapping model to align their distribution with the label distribution. Meanwhile, another encoding branch inputs the image into a learnable encoder, where consistency learning is applied to the unlabeled data using two distinct mapping strategies. The obtained features are then passed to the decoder, which includes two adapters (\ie, Mask Adapter and Image Adapter). Finally, the features are processed by the reverse transformation to generate the predicted segmentation maps.



\subsection{Dual-distribution Alignment Module}

\label{sec:DAM}

Traditional discriminative segmentation methods rely on large labeled datasets and optimize models through maximum likelihood estimation. However, these methods are generally ineffective at capturing global image structures and contextual semantics, particularly when applied to complex or annotation-limited medical scenes. In contrast, generative segmentation enhances performance by aligning latent spaces, where the model learns the posterior distribution of latent variables, thereby capturing underlying structural information and improving the understanding of intricate image details and fine-grained semantics. 

Unlike prior decoder-multiplexing paradigms, we utilize two distinct encoders to model the feature distribution transformation from images to masks. 
Specifically, the input image $x \in \mathbb{R}^{H \times W \times 3}$ is first processed through a pre-trained VAE~\cite{rombach2022high} encoder $\mathcal{E}(\cdot)$, to capture the prior distribution $p(z_v | x)$ of the image, which can be expressed as follows: 
\begin{equation}
p(z_v | x) = \mathcal{E}(x) := \mathcal{N}(z_v; \mu_{z_v}(x; \theta_v), \sigma_{z_v}(x; \theta_v)),
\end{equation}
where $\mu_{z_v}(x; \theta_v)$ and $\sigma_{z_v}(x; \theta_v)$ are the mean and standard deviation of the Gaussian distribution, respectively, parameterized by the VAE encoder with parameters $\theta_v$.
We then utilize a latent mapping model $\mathcal{M}(\cdot)$ that leverages the self-attention mechanism to map image features to a low-dimensional latent space, effectively capturing the global dependencies between the images and masks, expressed by
\begin{equation}
p(\tilde{z}_v | z_v) = \mathcal{M}(z_v) := \mathcal{N}(\tilde{z}_v; \mu_{\tilde{z}_v}(z_v; \theta_m), \sigma_{\tilde{z}_v}(z_v; \theta_m)),
\end{equation}
where $\mu_{\tilde{z}_v}$ and $\sigma_{\tilde{z}_v}$ are the mean and standard deviation of the Gaussian distribution, parameterized by the learnable parameters $\theta_m$ of the latent mapping model and dependent on $z_v$. 
To leverage unlabeled data and explore inherent data patterns through consistency learning, we employ two encoders with significant structural differences to provide diverse feature perspectives. Therefore, the same input image $x \in \mathbb{R}^{H \times W \times 3}$ is also fed into a trainable encoder ${E}(\cdot)$ (as shown in Fig.~\ref{fig:Model}), which focuses on extracting fine-grained structural and discriminative features. The prior distribution $p(z_r | x)$ is expressed as follows:
\begin{equation}
p(z_r | x) = {E}(x) := \mathcal{N}(z_r; \mu_{z_r}(x; \theta_r), \sigma_{z_r}(x; \theta_r)),
\end{equation}
where $\mu_{z_r}(x; \theta_r)$ and $\sigma_{z_r}(x; \theta_r)$ are the mean and standard deviation of the Gaussian distribution, respectively, parameterized by the learnable parameters $\theta_r$ of the trainable encoder ${E}(\cdot)$.
To constrain the prior distribution of segmentation masks with image feature distributions, we employ the VAE encoder $\mathcal{E}$ for mask processing, and for a ground-truth mask $g \in \mathbb{R}^{H \times W}$, only available for labeled data, its prior distribution $p(z_g | g)$ is modeled as:
\begin{equation}
p(z_g | g) = \mathcal{E}(g) := \mathcal{N}(z_g; \mu_{z_g}(g; \theta_v), \sigma_{z_g}(g; \theta_v)),
\end{equation}
where $\mu_{z_g}$ and $\sigma_{z_g}$ are the mean and standard deviation, parameterized by the VAE encoder's parameters $\theta_v$. 

To ensure consistency between the two-branch feature distributions and the mask prior in the latent space, as well as enhance branch alignment, we design a prior distribution alignment loss. Feature-level supervision, which is more informative than mask-level supervision, employs MSE to quantify the discrepancies among the distributions $p(\tilde{z}_v | z_v)$, $p(z_r | x)$, and the mask prior distribution $p(z_g | g)$. For labeled data, the loss is formulated by
\begin{equation}
\mathcal{L}_{sup}^{p} = \|\tilde{z}_v^l - {z_g}\|_2^2 + \|{z}_r^l - {z_g}\|_2^2,
\end{equation}
where ${\tilde{z}}_v^l$ and ${z}_r^l$ denote the two image branches' feature distributions under labeled data, respectively. Moreover, for unlabeled data, consistency learning is employed to constrain the distribution loss, which is expressed as follows:
\begin{equation}
\mathcal{L}_{unsup}^{p} = \|{\tilde{z}}_v^u - {z}_r^u\|_2^2,
\end{equation}
where ${\tilde{z}}_v^u$ and ${\tilde{z}}_v^u$ denote the feature distributions of the two image branches for unlabeled data, respectively.

\subsection{Consistency-Driven Skip Adapter}
\label{sec:CDSA}

In the segmentation task, while prior distribution alignment achieves image-mask feature-level consistency, gaps persist in multi-scale feature utilization and task-specific structural adaptation. Conventional VAE-based pipelines, despite effective distribution matching, often neglect fine-grained multi-scale details critical for segmentation. Moreover, image-centric and mask-centric feature distributions inherently differ in semantic granularity and domain specificity. To this end, we propose the Consistency-Driven Skip Adapter, which fuses multi-scale features via adapter-based skip connections to enhance cross-branch consistency.

\begin{figure}[!t]
	\centering
	\begin{overpic}[width=0.48\textwidth]{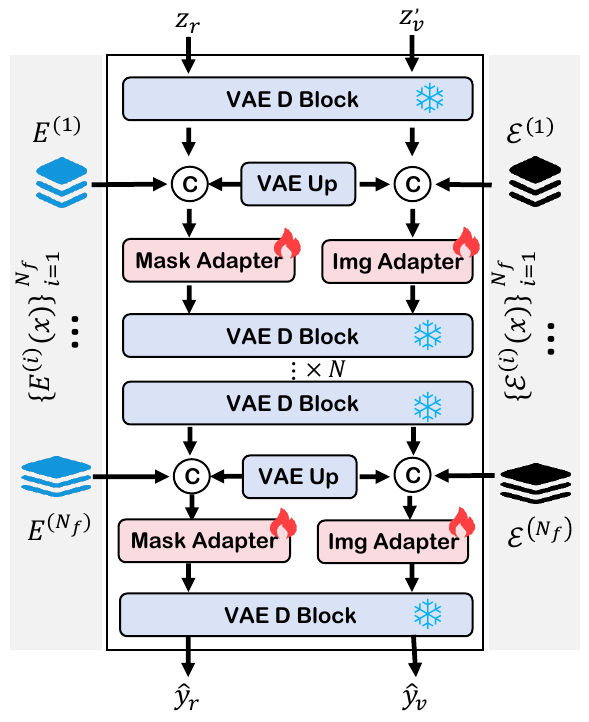}
    \end{overpic}\vspace{-0.15cm}
	\caption{{Structure of the skip connection adapter. ``VAE D Block'' and ``VAE Up'' represent the frozen VAE decoding block and upsampling block (denoted by blue blocks), where ``Img Adapter" and ``Mask Adapter" employ lightweight convolutional layers to efficiently adapt multi-scale skip connections.
 }}\vspace{-0.15cm}
    \label{fig:Decoder}
\end{figure}

Specifically, the module consists of two parallel components: an Image Skip Adapter and a Mask Skip Adapter, tailored for image distribution and mask distributions, respectively. As shown in \cref{fig:Decoder}, 
each adapter takes as input the hierarchical multi-scale features from its corresponding encoder at the skip connection points. We define the resulting two feature banks as follows:
\begin{equation}
S_v = \left\{ \mathcal{E}^{(i)} (x) \right\}_{i=1}^{N_f}, \quad 
S_r = \left\{ E^{(i)} (x)\right\}_{i=1}^{N_f},
\end{equation}
where ${N_f}$ denotes the number of multi-scale levels, $S_v$ represents the hierarchical multi-scale feature bank from the image-distribution encoder, and $S_r$ corresponds to that from the mask-distribution encoder. Note that the adapters are named based on the intrinsic differences between their feature distributions. Specifically, the Image Skip Adapter processes outputs from $\mathcal{E}^{(i)}(x)$, which retains image-distribution characteristics as VAE-compressed features. Conversely, the Mask Skip Adapter processes the outputs from $E^{(i)}(x)$, which inherits mask-distribution attributes through direct constraint by the label feature distribution. 


To enhance the intrinsic consistency of cross-branch features, we employ a consistency loss to enforce distribution alignment between the two adapters' outputs, ensuring their feature representations converge to a unified semantic space. For this constraint, we adopt the Dice loss, formally defined as follows:
\begin{equation}
\mathcal{L}_{dice}(\hat{y}, y) = 1 - \frac{2 \cdot \sum (\hat{y} \odot y)}{|\hat{y}|_1 + |y|_1},
\end{equation}
where $\hat{y}$ and $y$ respectively denote the prediction and ground-truth. 
For labeled data, each adapter's output is supervised by the ground-truth, which can be expressed by
\begin{equation}
\mathcal{L}_{sup}^{s} =  \mathcal{L}_{dice}(\hat{y}_v^l, y) + \mathcal{L}_{dice}(\hat{y}_r^l, y),
\end{equation}
where $\hat{y}_v^l$ and $\hat{y}_r^l$ represent the final predictions from the Image Skip Adapter and Mask Skip Adapter, respectively. 
For unlabeled data, the consistency constraint is imposed directly between the outputs of the two adapters, and the cross-branch consistency loss is formulated as follows:
\begin{equation}
\mathcal{L}_{unsup}^{s} =  \mathcal{L}_{dice}(\hat{y}_v^u, \hat{y}_r^u) + \mathcal{L}_{dice}(\hat{y}_r^u. \hat{y}_v^u).
\end{equation}
where $\hat{y}_v^u$ and $\hat{y}_r^u$ denote the predictions for the unlabeled data, respectively.
This paradigm, tailored to segmentation's demand for multi-scale and spatially coherent features, preserves fine-grained details and coarse-grained semantics at low computational cost and reinforces dual-branch consistency on unlabeled samples.

\subsection{Annotation Conversion and Reversion}
\label{sec:ACR}

We present an Annotation Conversion and Reversion (ACR) strategy to preprocess ground-truth (GT) annotations for compatibility with the VAE input in the segmentation paradigm. 
Specifically, for the input GT, we first normalize the original pixel values to the range [0, 1] by dividing them by the total number of classes \( K \), then map the result to [-1, 1] to meet the annotation input requirements of the VAE. The complete processing can be expressed as:  
\begin{equation}
G' = 2 \cdot \frac{G}{K} - 1,
\end{equation}  
where $G$ denotes the raw segmentation mask (with pixel values ranging from 0 to $K-1$ for $K$-class tasks), and $G'$ represents the processed GT fed into the VAE. This two-step transformation (normalization and range mapping) ensures compatibility with the input distribution characteristics of the VAE.
For the reversion step, the adopted end-to-end inverse calculation directly mirrors the forward conversion process, enabling full reversibility between the two operations.It not only guarantees strict consistency between the mask processing and reversion workflows but also preserves the semantic integrity of the segmentation mask throughout the pipeline, supporting reliable supervised training.

\begin{table*}[t!]
  \centering
  \scriptsize
  \small
 \renewcommand{\arraystretch}{1.0}
   \setlength\tabcolsep{1.7pt}
  \caption{Quantitative results on the colonoscopy and ISIC-2018 datasets. 
  }\vspace{-0.2cm}
  \label{tab:tab1}
  \begin{tabular}{r l|c|ccc|ccc|c|ccc|ccc}
  \hline
  \multicolumn{2}{c|}{\multirow{2}*{\textbf{Methods}}}  & & \multicolumn{3}{c|}{$10\%$ labeled} & \multicolumn{3}{c|}{$30\%$ labeled} &  & \multicolumn{3}{c|}{$10\%$ labeled} & \multicolumn{3}{c}{$30\%$ labeled}\\
  \cline{4-9}\cline{11-16}
   & & & Dice (\%) & IoU (\%)  & 95HD & Dice (\%) & IoU (\%) & 95HD & & Dice (\%) & IoU (\%)  & 95HD & Dice (\%) & IoU (\%) & 95HD\\
    \hline
  UA-MT \cite{yu2019uncertainty} &{\fontsize{6pt}{6pt}\selectfont\textcolor{gray}{[MICCAI'19]}} & \multirow{14}{*}{\begin{sideways}CVC-300\end{sideways}}  & 40.15 & 32.46 & 4.97 & 80.95 & 70.77 & {3.19}  & \multirow{14}{*}{\begin{sideways}CVC-ClinicDB\end{sideways}}  & 74.51 & 67.14 & 4.05 & 78.89 & 72.61 & 3.68\\
  DTC~\cite{luo2021semi} &{\fontsize{6pt}{6pt}\selectfont\textcolor{gray}{[AAAI'21]}} & & 42.81 & 33.35 & 5.42 & 77.44 & 67.43 & 3.32  & & 67.82 & 57.96 & 4.61 & 74.28 & 66.21 & 3.97\\
    MC-Net~\cite{wu2021semi}&{\fontsize{6pt}{6pt}\selectfont\textcolor{gray}{[MICCAI'21]}}  & & 69.89 & 61.58 & 3.93 & 81.36 & 72.11 & 3.33  & & 75.21 & 67.87 & 4.10 & 83.53 & 76.78 & 3.56\\
  URPC~\cite{luo2022semi} &{\fontsize{6pt}{6pt}\selectfont\textcolor{gray}{[MIA'22]}} & & 58.83 & 50.81 & 4.30 & 77.05 & 66.75 & 3.65  & & 76.42 & 69.33 & 3.78 & 80.70 & 74.58 & 3.64\\
  MCF~\cite{wang2023mcf} &{\fontsize{6pt}{6pt}\selectfont\textcolor{gray}{[CVPR'23]}} & & 67.58 & 57.85 & 3.91 & 79.51 & 69.03 & 3.27  & & 70.83 & 62.41 & 4.06 & 80.41 & 74.50 & 3.63\\
  CauSSL~\cite{miao2023caussl}&{\fontsize{6pt}{6pt}\selectfont\textcolor{gray}{[ICCV'23]}} & & 64.93 & 55.06 & 4.33 & 71.92 & 62.02 & 3.71  & & 76.23 & 68.44 & 3.99 & 75.18 & 68.08 & 3.89\\
  CDMA~\cite{zhong2023semi}&{\fontsize{6pt}{6pt}\selectfont\textcolor{gray}{[MICCAI'23]}} & & 59.44 & 48.93 & 4.34 & 66.38 & 52.84 & 4.77  & & 74.18 & 65.31 & 4.09 & 80.81 & 72.79 & 3.81\\
  BS-Net~\cite{he2023bilateral}&{\fontsize{6pt}{6pt}\selectfont\textcolor{gray}{[TMI'24]}} & & 65.46 & 56.31 & 4.09 & 80.53 & 70.99 & 3.33  & & 74.83 & 67.44 & 4.03 & 80.32 & 74.15 & 3.62\\
  PMT~\cite{gao2024pmtprogressivemeanteacher}&{\fontsize{6pt}{6pt}\selectfont\textcolor{gray}{[ECCV'24]}} & & 59.94 & 51.16 & 4.02 & 80.58 & 72.08 & 3.15  & & 72.17 & 64.62 & 3.76 & 74.35 & 67.17 & 3.55\\
  VCLIPSeg~\cite{li2024vclipseg}&{\fontsize{6pt}{6pt}\selectfont\textcolor{gray}{[MICCAI'24]}} & & 69.14 & 61.29 & {3.61} & 78.66 & 70.80 & 3.14  & & 76.48 & 68.85 & {3.70} & {81.61} & {75.37} & \textbf{3.36}\\
  UnCo~\cite{zeng2025uncertainty} &{\fontsize{6pt}{6pt}\selectfont\textcolor{gray}{[TMI'25]}} & & 77.56 & 69.64& 3.97 & 82.02 & 73.84 & 3.12  & & 78.01& 71.61 & 3.39 & 79.29& 72.45& 3.58\\
  SKCDF~\cite{zhang2025semantic}&{\fontsize{6pt}{6pt}\selectfont\textcolor{gray}{[CVPR'25]}} & & 57.28 & 48.24 & 4.03 & 75.68 & 64.09 & 3.62  & & 70.70 & 63.25 & 3.67 & 78.31 & 71.11 & 3.62\\
  CSCPA~\cite{ding2025csc}&{\fontsize{6pt}{6pt}\selectfont\textcolor{gray}{[CVPR'25]}} & & 76.97 & 65.99 & 3.57  & 81.54 & 70.97 & 3.08 && 78.75 & 70.64 & \textbf{3.38}  & 79.66 & 70.75 & 4.34\\
\hline

\textbf{Ours}&\rule[0.5ex]{2em}{0.4pt} & & \textbf{84.34} & \textbf{76.28} & \textbf{3.19}  & \textbf{86.75} & \textbf{79.26} & \textbf{3.07} & & \textbf{79.37} & \textbf{72.10} & {3.88}&   \textbf{83.88} & \textbf{77.22} & {3.56}\\

  \hline
  UA-MT \cite{yu2019uncertainty}&{\fontsize{6pt}{6pt}\selectfont\textcolor{gray}{[MICCAI'19]}} & \multirow{14}{*}{\begin{sideways}Kvasir\end{sideways}}  & 73.15 & 64.12 & 5.06 & 83.71 & 75.95 & 4.51  & \multirow{14}{*}{\begin{sideways}ISIC-2018\end{sideways}} & 80.15 & 71.48 & 5.37 & 83.86 & 75.54 & 4.81\\
  DTC~\cite{luo2021semi}&{\fontsize{6pt}{6pt}\selectfont\textcolor{gray}{[AAAI'21]}} & & 66.79 & 56.00 & 5.63 & 77.73 & 68.78 & 4.73  & & 80.02 & 71.74 & 5.19 & 83.53 & 75.21 & 4.89\\
    MC-Net~\cite{wu2021semi} & {\fontsize{6pt}{6pt}\selectfont\textcolor{gray}{[MICCAI'21]}}& & 78.03 & 69.21 & 4.89 & 85.51 & 78.07 & 4.46  & & 80.98 & 72.03 & 5.31 & 83.21 & 74.42 & 4.98\\
  URPC~\cite{luo2022semi}& {\fontsize{6pt}{6pt}\selectfont\textcolor{gray}{[MIA'22]}}& & 79.14 & 70.53 & 4.65 & 84.47 & 76.63 & 4.34  & & 79.38 & 70.36 & 5.37 & 83.18 & 74.85 & 4.91\\
  MCF~\cite{wang2023mcf}&{\fontsize{6pt}{6pt}\selectfont\textcolor{gray}{[CVPR'23]}} & & 74.01 & 64.65 & 4.99 & 83.25 & 75.42 & 4.46  & & 80.76 & 71.88 & 5.09 & 83.50 & 75.13 & 4.86\\
  CauSSL~\cite{miao2023caussl}&{\fontsize{6pt}{6pt}\selectfont\textcolor{gray}{[ICCV'23]}} & & 76.48 & 67.61 & 4.93 & 81.03 & 73.20 & 4.67  & & 79.73 & 71.19 & 5.34 & 84.53 & 75.88 & 4.93\\
  CDMA~\cite{zhong2023semi}&{\fontsize{6pt}{6pt}\selectfont\textcolor{gray}{[MICCAI'23]}} & & 74.98 & 65.06 & 5.07 & 82.43 & 73.47 & 4.63  & & 76.70 & 66.58 & 5.84 & 83.27 & 74.88 & 4.78\\
  BS-Net~\cite{he2023bilateral}& {\fontsize{6pt}{6pt}\selectfont\textcolor{gray}{[TMI'24]}}& & 77.18 & 68.22 & 4.99 & 82.50 & 74.98 & 4.45  & & 81.70 & 73.68 & 5.05 & 84.22 & 75.83 & 4.85\\
  PMT~\cite{gao2024pmtprogressivemeanteacher}&{\fontsize{6pt}{6pt}\selectfont\textcolor{gray}{[ECCV'24]}} & & 75.16 & 66.18 & {4.49} & 81.05 & 73.16 & 4.35  & & 82.37 & 74.18 & 4.80 & 84.99 & 76.73 & 4.59\\
  VCLIPSeg~\cite{li2024vclipseg}& {\fontsize{6pt}{6pt}\selectfont\textcolor{gray}{[MICCAI'24]}}& & 77.22 & 67.75 & 4.96 & 83.79 & 76.01 & 4.26  & & 79.37 & 70.32 & 5.39 & 83.60 & 74.91 & 4.99\\
  UnCo~\cite{zeng2025uncertainty} &{\fontsize{6pt}{6pt}\selectfont\textcolor{gray}{[TMI'25]}} & & 81.19 & 73.79 & 4.60 & 85.04 & 77.99 & 4.24  & & 85.59 & 77.56 & 4.77 & 87.06 & 79.59 & 4.62\\
  SKCDF~\cite{zhang2025semantic}&{\fontsize{6pt}{6pt}\selectfont\textcolor{gray}{[CVPR'25]}} & & 74.14 & 64.43 & 5.11 & 81.72 & 73.57 & 4.56  & & 84.86 & 77.45 & 4.74 & 86.06 & 77.92 & 4.59\\
  CSCPA~\cite{ding2025csc}&{\fontsize{6pt}{6pt}\selectfont\textcolor{gray}{[CVPR'25]}} & & 81.60 & 72.14 & 4.54 & 87.31 & 78.56 & 4.08  & & 85.75 & 78.25 & 4.81 & 86.47 & 78.48 & 4.59\\

\hline
  \textbf{Ours}&\rule[0.5ex]{2em}{0.4pt} & &\textbf{83.03} & \textbf{72.91} & \textbf{4.39} & \textbf{89.01} & \textbf{80.69} & \textbf{3.99}  & & \textbf{86.28} & \textbf{78.47} &  \textbf{4.70} & \textbf{87.62} & \textbf{80.51} & \textbf{4.53}\\

  \hline
  \end{tabular}\vspace{-0.1cm}
\end{table*}

\subsection{Overall Loss Function}\label{sec:loss}


The overall loss function of the framework is composed of supervised and unsupervised components
Specifically, the supervised loss uses labels as supervision signals, including latent space prior distribution loss~(described in \cref{sec:DAM}) to constrain the distribution of latent variables and segmentation loss~(described in \cref{sec:CDSA}) to measure the discrepancy with the ground truth. The overall expression is:
\begin{equation}
\mathcal{L}_{sup} = \mathcal{L}_{sup}^p + \mathcal{L}_{sup}^s.
\end{equation}  

For unlabeled data, model optimization relies on unsupervised consistency constraints. This is accomplished by enforcing output consistency and distributional consistency loss, which can be formulated as:
\begin{equation}
\mathcal{L}_{unsup} = \mathcal{L}_{unsup}^p + \mathcal{L}_{unsup}^s.
\end{equation} 

Finally, the total loss function of the framework is the sum of the supervised and unsupervised components, which is formulated as follows:  
\begin{equation}
\mathcal{L}_{total} = \mathcal{L}_{sup} + \lambda_u \mathcal{L}_{unsup},
\end{equation}  
where $\lambda_u$ is governed by a Gaussian warm-up schedule defined as $\lambda_{u}(t) = \beta \cdot e^{-5(1-t/t_{max})^2}$, where $\beta$ is set to 0.1 and $t_{max}$ represents the total number of training iterations.

\section{Experiments}

\begin{figure*}[!t]
	\centering
	\begin{overpic}[width=0.95\textwidth]{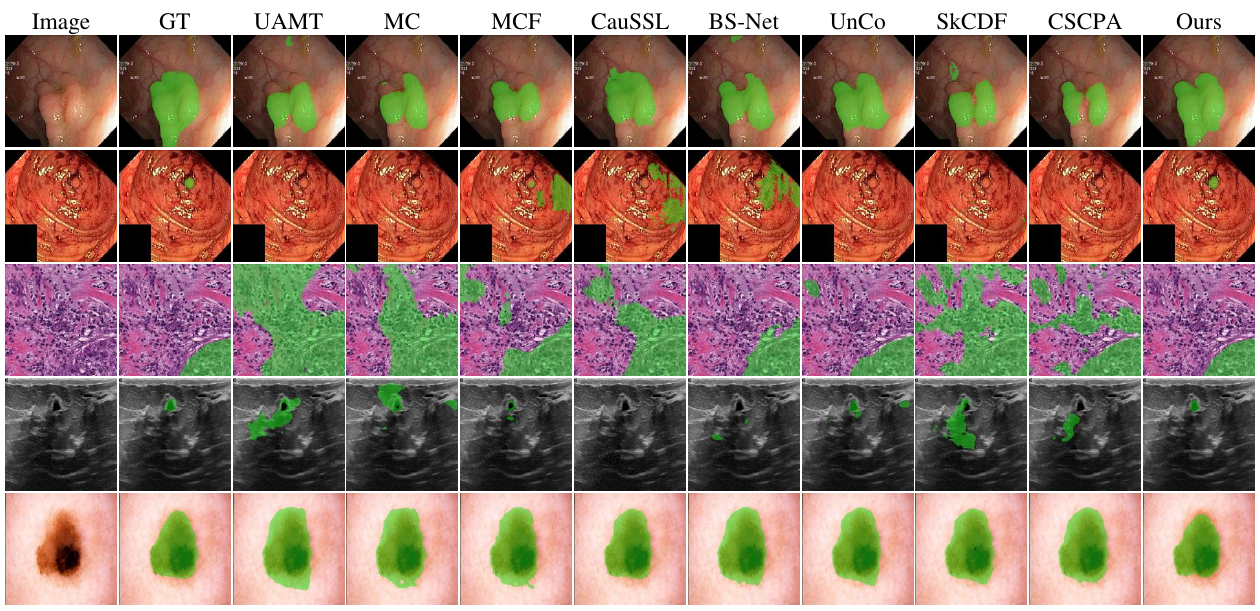}
    \end{overpic}\vspace{-0.1cm}
	\caption{{Visual comparisons of our model and other state-of-the-art semi-supervised medical segmentation methods. 
 }}\vspace{-0.2cm}
    \label{fig:showRes}
\end{figure*}

\subsection{Datasets}

We conduct experiments on four medical image segmentation tasks:
$\bullet$ \textbf{Colonoscopy}: We use CVC-ClinicDB~\cite{bernal2015wm}, Kvasir~\cite{jha2020kvasir}, and CVC-300~\cite{vazquez2017benchmark}. Following~\cite{fan2020pranet}, 1,450 images from ClinicDB and Kvasir form the training set (10\% for validation). The remaining images and the entirety of CVC-300 are used for testing.
$\bullet$ \noindent\textbf{ISIC-2018}: This dataset \cite{codella2019skin} contains an extensive collection of dermoscopy images,  with $2,075$ images for training, $519$ images for validation, and $1,000$ images for testing. 
$\bullet$ \noindent\textbf{BCSS}~\cite{amgad2019structured}: We crop 151 pathological images into 3,888 patches ($1024 \times 1024$), treating only tumor regions as foreground. The data is split into 70\% training, 10\% validation, and 20\% testing.
$\bullet$ \noindent\textbf{BUSI}~\cite{al2020dataset}: This dataset~\cite{al2020dataset} aggregates 780 images from 600 female participants with an average size of 500$\times$500 pixels. The samples include 133 normal, 210 malignant, and 437 benign cases. Our study utilizes the 647 abnormal samples, which are split into training and test subsets at 80\% and 20\%, respectively.

\subsection{Implementation Details}

All experiments are conducted on two NVIDIA 4090 GPUs using PyTorch 2.4.1 and CUDA 11.2 settings.
We use the Stable Diffusion (SD) VAE weights~\cite{rombach2022high} as pretrained encoder~$\mathcal{E}(\cdot)$  and decoder~$\mathcal{D}(\cdot)$ , leveraging their rich latent representation for strong zero-shot generalization. 
To better adapt the generative model to downstream segmentation tasks, we divide the training into two stages. We first pre-train the mapping network~$\mathcal{M}(\cdot)$ and the encoder~${E}(\cdot)$, and then we fine-tune the entire model in the second stage. This strategy enhances training stability. 
The batch size is set to $4$, including $2$ labeled and $2$ unlabeled samples. 
The input images are resized to $224 \times 224$. 
Moreover, the pretraining phase consists of $200$ epochs, followed by full training for $350$ epochs.
During the inference phase, we use the average of the two predictions as the final result.

\begin{table}[t!]
  \centering
  \scriptsize
  \renewcommand{\arraystretch}{1.1}
  \setlength\tabcolsep{1.6pt}
  \caption{Quantitative results on the BCSS and BUSI datasets.}\label{tab_BCSS_BUSI}\vspace{-0.15cm}
\begin{tabular}{r l|c c c|c c c}
\hline
{\multirow{2}{*}{Methods}} & & \multicolumn{3}{c|}{BCSS~(10\%)}    & \multicolumn{3}{c}{BCSS~(30\%)} \\
\cline{3-8}
&       & Dice (\%)  & IoU (\%)  & 95HD             & Dice (\%)  & IoU (\%)  & 95HD   \\
\hline
UA-MT~\cite{yu2019uncertainty} &{\fontsize{6pt}{6pt}\selectfont\textcolor{gray}{[MICCAI'19]}} & 63.05  & 49.65  &  9.28 & 67.00 & 54.28 & 8.81 \\
DTC~\cite{luo2021semi}   & {\fontsize{6pt}{6pt}\selectfont\textcolor{gray}{[AAAI'21]}}   &  63.60  & 50.19  &  9.36 & 69.44 & 57.07 & 8.32 \\
MC-Net~\cite{wu2021semi} & {\fontsize{6pt}{6pt}\selectfont\textcolor{gray}{[MICCAI'21]}}  &   63.67  & 50.12  &  9.21  & 69.55 & 57.33 & 8.36 \\
URPC~\cite{luo2022semi} &{\fontsize{6pt}{6pt}\selectfont\textcolor{gray}{[MIA'22]}} & 63.93  & 50.15  &  9.23 & 68.92 & 56.11 & 8.63 \\
MCF~\cite{wang2023mcf} &{\fontsize{6pt}{6pt}\selectfont\textcolor{gray}{[CVPR'23]}} &  65.52  & 52.14  &  8.98 & 70.49 & 57.91 & 8.30 \\
CDMA~\cite{zhong2023semi}& {\fontsize{6pt}{6pt}\selectfont\textcolor{gray}{[MICCAI'23]}} &  65.58  & 52.25  &  8.92 & 67.28 & 54.57 & 8.65 \\
CauSSL~\cite{miao2023caussl} &{\fontsize{6pt}{6pt}\selectfont\textcolor{gray}{[ICCV'23]}} & 62.88  & 49.35  &  9.45 & 71.12 & 58.69 & 8.32 \\
BS-Net~\cite{he2023bilateral}&{\fontsize{6pt}{6pt}\selectfont\textcolor{gray}{[TMI'24]}} &  64.47  & 52.37  &  {8.28}  & 71.30 & 58.95 & 8.05 \\
PMT~\cite{gao2024pmtprogressivemeanteacher} &{\fontsize{6pt}{6pt}\selectfont\textcolor{gray}{[ECCV'24]}}& 68.97 & 56.37 & 8.06 & 71.45 & 59.75 & 7.48 \\
VCLIPSeg~\cite{li2024vclipseg}& {\fontsize{6pt}{6pt}\selectfont\textcolor{gray}{[MICCAI'24]}}&  68.14 & 55.80 &  7.84 & 71.77 & 60.07 &  7.46 \\
UnCo~\cite{zeng2025uncertainty} &{\fontsize{6pt}{6pt}\selectfont\textcolor{gray}{[TMI'25]}}& 68.66 & 56.62 &  7.69 & 71.56 & 60.17 &  7.38 \\
SKCDF~\cite{zhang2025semantic}&{\fontsize{6pt}{6pt}\selectfont\textcolor{gray}{[CVPR'25]}}&  70.31 & 57.65 &  7.67 & 73.84 & 62.07 &  7.26 \\
CSCPA~\cite{ding2025csc}&{\fontsize{6pt}{6pt}\selectfont\textcolor{gray}{[CVPR'25]}}& 71.95 & 59.48 &  7.57 & 73.40 & 61.89 &  7.13   \\
\hline
\textbf{Ours} & \rule[0.5ex]{2em}{0.4pt} &  \textbf{74.05} & \textbf{62.68} &  \textbf{7.05} & \textbf{75.65} & \textbf{64.59} & \textbf{6.90}  \\
\hline
\hline
& & \multicolumn{3}{c|}{BUSI~(10\%)}    & \multicolumn{3}{c}{BUSI~(30\%)} \\
\hline
UA-MT~\cite{yu2019uncertainty}&{\fontsize{6pt}{6pt}\selectfont\textcolor{gray}{[MICCAI'19]}}   & 57.85  & 49.07  & 4.96  & 67.82 & 58.66 & 4.67 \\
DTC~\cite{luo2021semi}  & {\fontsize{6pt}{6pt}\selectfont\textcolor{gray}{[AAAI'21]}}   &  52.73 & 42.69 & 5.26  & 63.67 & 53.57 & 4.76 \\
MC-Net~\cite{wu2021semi} & {\fontsize{6pt}{6pt}\selectfont\textcolor{gray}{[MICCAI'21]}} & 61.72  & 52.77  & 4.42  & 69.61 & 60.13 & 4.92 \\
URPC~\cite{luo2022semi} &{\fontsize{6pt}{6pt}\selectfont\textcolor{gray}{[MIA'22]}} & 57.29 & 48.64 & 4.54 & 64.65 & 53.10 & 5.78 \\
MCF~\cite{wang2023mcf} &{\fontsize{6pt}{6pt}\selectfont\textcolor{gray}{[CVPR'23]}} &  64.13 & 54.57  & 4.95  & 73.24 & 64.68 & 4.57 \\
CDMA~\cite{zhong2023semi} & {\fontsize{6pt}{6pt}\selectfont\textcolor{gray}{[MICCAI'23]}} &62.32  & 51.26  & 5.20 & 73.78 & 64.07 & 4.70 \\
CauSSL~\cite{miao2023caussl} &{\fontsize{6pt}{6pt}\selectfont\textcolor{gray}{[ICCV'23]}} & 58.12  & 47.87  & 5.07& 65.94 & 57.13 & 4.68 \\
BS-Net~\cite{he2023bilateral} &{\fontsize{6pt}{6pt}\selectfont\textcolor{gray}{[TMI'24]}} & 63.01  & 52.48  & 5.24 & 72.28 & 63.60 & 4.56 \\
PMT~\cite{gao2024pmtprogressivemeanteacher} &{\fontsize{6pt}{6pt}\selectfont\textcolor{gray}{[ECCV'24]}}& 60.49 & 52.63 & 3.91 & 69.48 & 61.93 & 4.63 \\
VCLIPSeg~\cite{li2024vclipseg}& {\fontsize{6pt}{6pt}\selectfont\textcolor{gray}{[MICCAI'24]}}& 59.01 & 49.68 &  4.90 & 72.09 & 63.45 &  4.52 \\
UnCo~\cite{zeng2025uncertainty} &{\fontsize{6pt}{6pt}\selectfont\textcolor{gray}{[TMI'25]}}& 62.24 & 54.42&  4.27  & 70.24 & 62.83 &  4.66 \\
SKCDF~\cite{zhang2025semantic}&{\fontsize{6pt}{6pt}\selectfont\textcolor{gray}{[CVPR'25]}}&  60.99 & 50.68 &  5.16 & 72.91 & 64.04 &  4.74 \\
CSCPA~\cite{ding2025csc}&{\fontsize{6pt}{6pt}\selectfont\textcolor{gray}{[CVPR'25]}}& 65.16 & 55.17 &  4.87   & 73.53 & 65.05 &  4.61   \\
\hline
\textbf{Ours} & \rule[0.5ex]{2em}{0.4pt}  &   \textbf{75.57} & \textbf{65.72} & \textbf{4.70}  & \textbf{78.74} & \textbf{69.49} & \textbf{4.52}  \\
  \hline
  \end{tabular}\vspace{-0.15cm}
\end{table}

\subsection{Comparison with State-of-the-art Methods}

We compare the proposed model with 11 SOTA semi-supervised medical image segmentation methods. To evaluate these methods, we employ three commonly used metrics, namely the Dice coefficient (Dice), Intersection over Union (IoU), and 95\% Hausdorff Distance (95HD). 

\textbf{Results on Colonoscopy and ISIC Datasets:} \cref{tab:tab1} presents quantitative results on colonoscopy datasets (CVC-300, CVC-ClinicDB) and ISIC-2018 under $10\%$ and $30\%$ labeled data. It can be observed that our model achieves promising performance than other comparison methods. 
For example, on Kvasir with $10\%$ labels, our method enhances Dice by $1.84\%$ and $8.89\%$ compared to UnCo and SKCDF, respectively. On ISIC-2018 with $10\%$ labels, compared to CSCPA, our method increases Dice from $85.75\%$ to $86.28\%$ and decreases 95HD from $4.81$ to $4.70$. As shown in \cref{fig:showRes}, our model segments target regions with more precise boundaries and improved completeness compared to methods like UA-MT and MC. 

\textbf{Results on BCSS and BUSI:} \cref{tab_BCSS_BUSI} presents quantitative results on the BCSS (pathology) and BUSI (ultrasound) datasets. 
Our model achieves leading performance across various labeling ratios on two distinct domains. 
On the BCSS dataset with $10\%$ labeled data, our model achieves $74.05\%$ Dice and $62.68\%$ IoU, surpassing the second-best method~(CSCPA) in terms of Dice and IoU by $2.10\%$ and $3.20\%$, respectively. 
This superiority is consistent on the BUSI dataset, where our model shows significant gains in both segmentation accuracy and boundary precision. As shown in \cref{fig:showRes}, our model achieves more precise segmentation in complex regions compared to other methods.

\subsection{Ablation Study}

\begin{figure}[!t]
	\centering
	\begin{overpic}
    [width=0.5\textwidth]{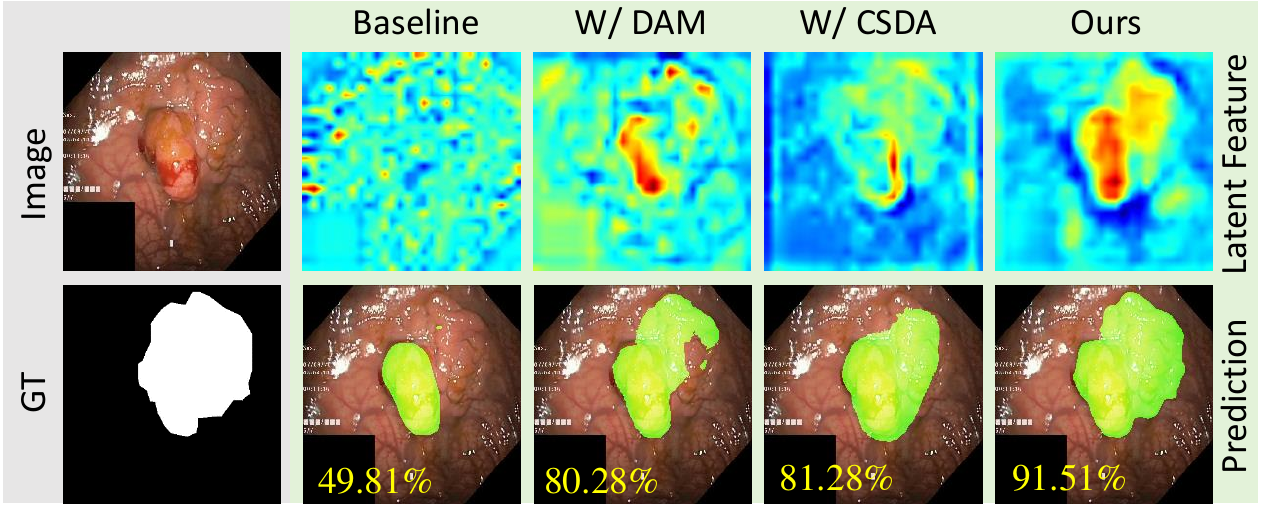}
    \end{overpic}
    \vspace{-0.6cm}
	\caption{{Visual maps of latent features and segmentation results. Yellow numbers denote the sample-level Dice scores. }}
    \vspace{-0.2cm}
    \label{fig:latent_features_vis}
\end{figure}

\begin{figure}[!t]
	\centering
	\begin{overpic}[width=0.48\textwidth]{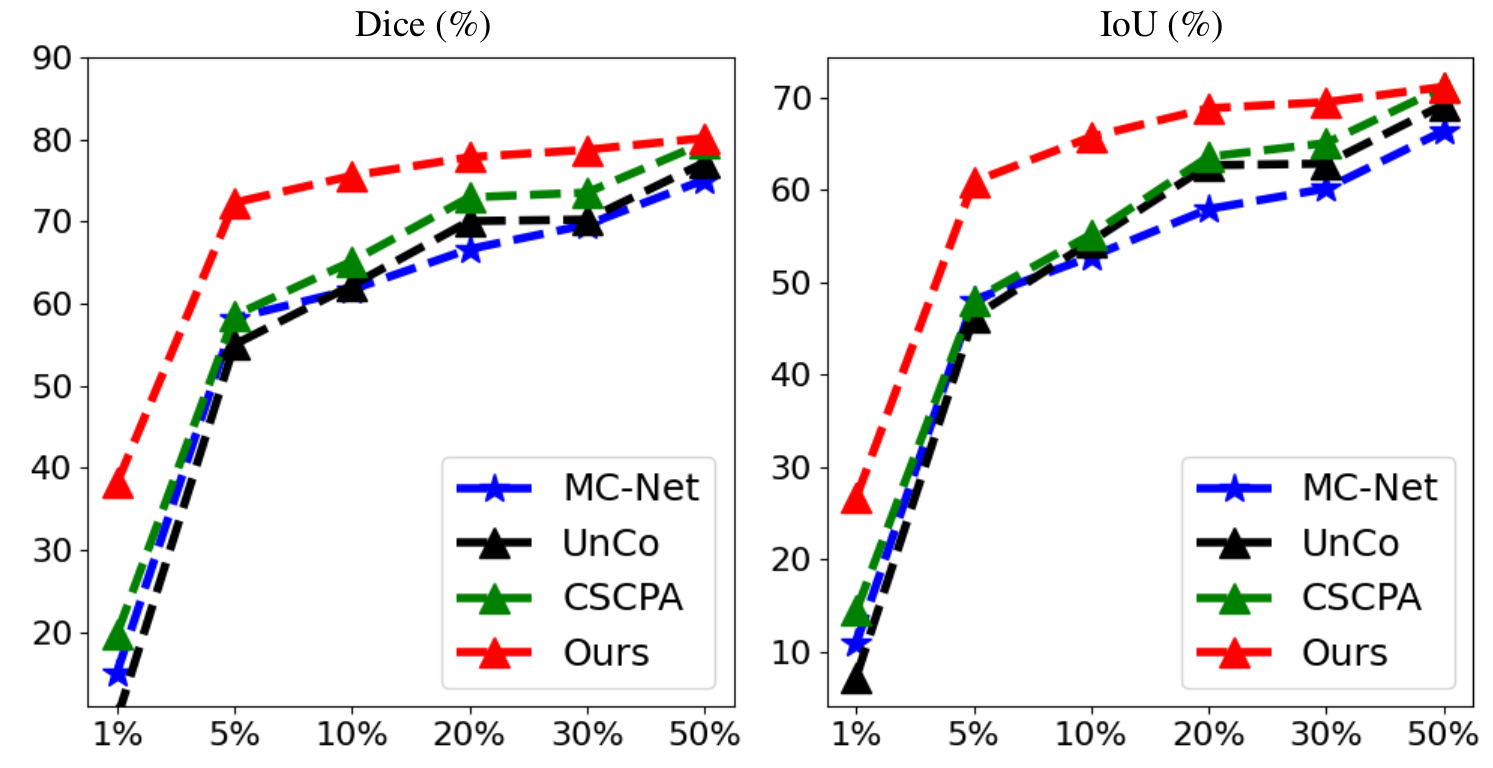}
    \end{overpic}\vspace{-0.15cm}
	\caption{{Performance comparison with different labeled ratios on the BUSI dataset.
 }}\vspace{-0.25cm}
    \label{fig:line}
\end{figure}

\textbf{Effects of Key Components}. 
Given the specific nature of our generative framework, traditional semi-supervised architectures like UAMT~\cite{yu2019uncertainty} or MC~\cite{wu2021semi} are not directly applicable. We therefore establish a baseline model by removing the feature-layer constraints and skip adapter, retaining only the final mask supervision. We then sequentially integrate our two core components: the DAM and the CDSA. 
As shown in~\cref{tab:tab_abi_component}, each component delivers significant performance gains. For example, on the BUSI dataset with $10\%$ labeled data, the Dice coefficient improves from $70.48\%$ for the baseline to $73.07\%$ with DAM, and further to $75.25\%$ with the addition of the CDSA.
Moreover, we provide a step-wise visualization in \cref{fig:latent_features_vis}. It shows that while the features of \textit{baseline (Col. 2)}  are scattered and noisy, the proposed \textit{DAM (Col. 3)} effectively rectifies the distribution to focus on the lesion, and \textit{CSDA (Col. 4)} further sharpens the boundaries.

\textbf{Effects of Two Skip Adapters in CDSA}. 
To investigate the effectiveness of skip adapters in applying generative models to segmentation, we conduct an ablation study in which the baseline retains all settings except for the removal of the two skip adapters. 
As shown in~\cref{tab_abi_skip}, on BUSI with $10\%$ labeled data, compared to the baseline, the Dice coefficient improves from $73.07\%$ to $75.57\%$, and the 95HD decreases from $5.30$ to $4.70$. This confirms that skip adapters are crucial for enhancing segmentation accuracy and improving the utilization of sparse labeled data. 

\textbf{Effects of Loss Functions}. 
To evaluate the loss function, our baseline retains all structural components (\eg~skip adapters) but excludes the unnecessary loss. As shown in \cref{tab_abi_loss}, incorporating the unsupervised loss significantly boosts the Dice score (\eg~from $79.34\%$ to $83.03\%$ on Kvasir with $10\%$ labels). This confirms that our generative approach enhances pseudo-label quality, demonstrating strong robustness in leveraging unlabeled data.

\textbf{Performance under Different Labeled Ratios}. We further evaluate the performance under different labeled ratios. As shown in~\cref{fig:line}, our model consistently outperforms both representative (\eg, MC-Net) and state-of-the-art (\eg, UnCo and CSCPA) approaches. Notably, our model excels with extremely limited labeled data (\ie, $1\%$), underscoring the superior robustness of generative segmentation over discriminative approaches in low-label scenarios.


\begin{table}[t!]
  \centering
  \footnotesize
  \renewcommand{\arraystretch}{1.0}
  \setlength
  \tabcolsep{1.0pt}
  \caption{Ablation study on each key component in our method.} \vspace{-0.1cm}
  \label{tab:tab_abi_component}
\begin{tabular}{c||cc|cc|cc|cc}
  \hline
\multirow{2}{*}{Base.} & \multicolumn{2}{c|}{GDA} & \multicolumn{2}{c|}{\cellcolor[gray]{0.9}ClinicDB~(10\%)} & \multicolumn{2}{c|}{\cellcolor[gray]{0.9}Kvasir~(10\%)} & \multicolumn{2}{c}{\cellcolor[gray]{0.9}BUSI~(10\%)} \\
\cline{2-9}
& DAM    & CDSA    & \cellcolor[gray]{0.9}Dice~(\%)  & \cellcolor[gray]{0.9}95HD   & \cellcolor[gray]{0.9}Dice~(\%) & \cellcolor[gray]{0.9}95HD & \cellcolor[gray]{0.9}Dice~(\%) & \cellcolor[gray]{0.9}95HD \\
\hline
\checkmark &     &       &    74.23   &  4.47     & 77.01       &    5.03    &  70.48 &  5.44   \\
\checkmark &   \checkmark   &    &    75.92 &  4.21    &    80.02   &  4.70       &    73.07    &   5.30    \\
\checkmark &     &   \checkmark    &    76.83   &  4.07    &  81.84       &    4.54   & 75.25       &    4.77 \\
\checkmark & \checkmark    & \checkmark   &    \textbf{79.37}   &  \textbf{3.88}     &\textbf{83.03}       &\textbf{4.39}    & \textbf{75.57}       &    \textbf{4.70}       \\
\hline
\end{tabular}
\vspace{-0.0cm}
\end{table}

\begin{table}[t!]
  \centering
  \footnotesize
  \renewcommand{\arraystretch}{1.0}
  \setlength
  \tabcolsep{2.8pt}
 \caption{Ablation study on the two skip adapters in CDSA.}\vspace{-0.2cm}
  \label{tab_abi_skip}
\begin{tabular}{cc||cc|cc|cc}
\hline
\multicolumn{2}{c||}{Settings} & \multicolumn{2}{c|}{\cellcolor[gray]{0.9}ClinicDB~(10\%)} & \multicolumn{2}{c|}{\cellcolor[gray]{0.9}Kvasir~(10\%)} & \multicolumn{2}{c}{\cellcolor[gray]{0.9}BUSI~(10\%)} \\
\hline
Image        & Mask      & \cellcolor[gray]{0.9}Dice~(\%)  & \cellcolor[gray]{0.9}95HD   & \cellcolor[gray]{0.9}Dice~(\%) & \cellcolor[gray]{0.9}95HD & \cellcolor[gray]{0.9}Dice~(\%) & \cellcolor[gray]{0.9}95HD     \\
\hline
 &           &    75.92 &  4.21     &    80.02        &  4.70     &    73.07    &   5.30   \\
   \checkmark &          & 76.98    &   4.25    &    81.95        &  4.69      &   74.92   &   4.75          \\
 &   \checkmark       & 77.14    &   4.11    &  81.79     &    4.52     &   73.91   &   4.88          \\
\checkmark &   \checkmark      &    \textbf{79.37}   &  \textbf{3.88}     &\textbf{83.03}       &\textbf{4.39}   & \textbf{75.57}       &    \textbf{4.70}       \\
\hline
\end{tabular}
\vspace{-0.05cm}
\end{table}

\begin{table}[t!]
  \centering
  \footnotesize
  \renewcommand{\arraystretch}{1.0}
  \setlength
  \tabcolsep{4.0pt}
  \caption{Ablation study on loss functions ($10\%$ labeled data).}\vspace{-0.1cm}
  \label{tab_abi_loss}
\begin{tabular}{c||cccc|cc}
\hline
\multirow{2}{*}{Datasets} & \multicolumn{4}{c|}{\cellcolor[gray]{0.9}Loss Functions}                  & \multicolumn{2}{c}{\cellcolor[gray]{0.9}Metrics} \\
\cline{2-7}
  & \cellcolor[gray]{0.9}$\mathcal{L}_{sup}^s$ & \cellcolor[gray]{0.9}$\mathcal{L}_{sup}^p$ & \cellcolor[gray]{0.9}$\mathcal{L}_{unsup}^s$  &\cellcolor[gray]{0.9} $\mathcal{L}_{unsup}^p$ & \cellcolor[gray]{0.9} Dice~(\%)      & \cellcolor[gray]{0.9}HD95    \\

\hline  
\multirow{4}{*}{ClinicDB}    
    &   \checkmark  &   \checkmark    &     &      &   78.61  &   4.97 \\
                    &   \checkmark  &   \checkmark    &    \checkmark     &    &   78.77       &   4.90  \\
                    &   \checkmark  &   \checkmark    &       &    \checkmark  &   78.80       &   4.78   \\
          &   \checkmark  &   \checkmark    &    \checkmark  &    \checkmark  &    \textbf{79.37}   &  \textbf{3.88}  \\
\hline  
\multirow{4}{*}{Kvasir}    
    &   \checkmark  &   \checkmark    &     &      &   79.34       &   4.63  \\
                    &   \checkmark  &   \checkmark    &    \checkmark     &    &   82.64       &   4.52  \\
                    &   \checkmark  &   \checkmark    &       &    \checkmark   &   81.10     &   4.57  \\
          &   \checkmark  &   \checkmark    &    \checkmark  &    \checkmark  &\textbf{83.03}       &\textbf{4.39} \\
\hline  
\multirow{4}{*}{BUSI}    
                 &   \checkmark  &   \checkmark    &     &      &    75.08   &  4.89 \\
                &   \checkmark  &   \checkmark    &    \checkmark     &    &  75.36     &   4.75  \\
                &   \checkmark  &   \checkmark    &       &    \checkmark  &  75.32      & 4.88   \\
      &   \checkmark  &   \checkmark    &    \checkmark  &    \checkmark   & \textbf{75.57} & \textbf{4.70}  \\
\hline  
\end{tabular}
\vspace{-0.0cm}
\end{table}


\section{Conclusion}

In this paper, we propose SemiGDA, a novel generative segmentation framework to address overfitting and poor pseudo-label quality in settings with limited labeled data. The framework consists of two key components: DAM and CDSA. DAM uses two learnable components to map the prior distributions from images to masks, improving model robustness through feature distribution-level supervision. CDSA introduces skip adapters into the decoder of a pre-trained generative model, enhancing its adaptation to domain-specific segmentation tasks. 
Additionally, we resolve the compatibility issue between ground truth and the generative model's input-output structure with a training-free conversion and reversion strategy. 
Experimental results on four segmentation tasks show that SemiGDA outperforms existing SOTA semi-supervised methods. 

\section{Acknowledgments}

This work was supported in part by the National Natural Science Foundation of China (Nos. 62576153, 62476054).

{
    \small
    \bibliographystyle{ieeenat_fullname}
    \bibliography{main}
}

\end{document}